\def\year{2019}\relax
\documentclass[letterpaper]{article} 
\usepackage{dstc8}  
\usepackage{amsfonts}
\usepackage{amsmath}
\usepackage{times}  
\usepackage{helvet} 
\usepackage{courier}  
\usepackage[hyphens]{url}  
\usepackage{graphicx} 
\usepackage{graphicx}  
\usepackage{tabularx}
\usepackage{multirow}
\usepackage[textwidth=10ex]{todonotes}
\usepackage{footnote}
\usepackage{booktabs}

\newcommand{\shortciteauthor}[1]{\citeauthor{#1} \shortcite{#1}}
\newcommand{\citesystem}[2]{(#1; \citeauthor{#2} \citeyear{#2})}

\title{Show Us the Way: Learning to Manage Dialog from Demonstrations}
\author{Gabriel Gordon-Hall\thanks{equal contribution}, Philip John Gorinski\footnotemark[1], Gerasimos Lampouras, Ignacio Iacobacci\\
	Huawei Noah's Ark Lab, London, UK\\
	\{gabriel.gordon.hall, philip.john.gorinski, gerasimos.lampouras, ignacio.iacobacci\}@huawei.com
}

\begin{document}
	
	\maketitle
	
	\begin{abstract}
		We present our submission to the End-to-End Multi-Domain Dialog Challenge Track of the Eighth Dialog System Technology Challenge. Our proposed dialog system adopts a \emph{pipeline} architecture, with distinct components for Natural Language Understanding, Dialog State Tracking, Dialog Management and Natural Language Generation. At the core of our system is a reinforcement learning algorithm which uses Deep Q-learning from Demonstrations to learn a dialog policy with the help of expert examples. We find that demonstrations are essential to training an accurate dialog policy where both state and action spaces are large. Evaluation of our Dialog Management component shows that our approach is effective - beating supervised and reinforcement learning baselines.
	\end{abstract}
	
	\section{Introduction}
	Conversational AI is an area of Natural Language Processing (NLP) that deals with the modelling of user-agent dialogues. Over the years, the Dialog System Technology Challenge (DSTC) has provided valuable research impulses for this community, and the release of large-scale multi-domain datasets like MultiWOZ \cite{budzianowski2018multiwoz,eric2019multiwoz} has enabled the development of high performing multi-domain dialog systems. Conversational AI can be broadly divided into three categories: task-oriented, question-answering, and chit-chat systems \cite{gao2019neural}. Chit-chat systems aim to mimic the flow of human conversation, and have no goal other than to keep the user engaged. Question-answering systems have the specific aim of answering user questions, often pertaining to information encoded in a knowledge-graph. Finally, the aim of task-oriented systems is the successful simulation of a dialog \emph{agent} to help users achieve a certain task, such as booking flights or hotels. More recently, simulations of \emph{users} have attracted attention\footnote{https://ai.googleblog.com/2018/05/duplex-ai-system-for-natural-conversation.html}, and some work has even tried to simulate both sides of the full dialog \cite{das2017learning}.
	
	At the highest level, task-oriented systems can be grouped into ``monolithic'' and ``pipelined'' approaches. In the latter group, the overall problem is broken down into smaller components, with systems typically having separate modules for Natural Language Understanding (NLU), Dialog State Tracking (DST), Dialog Management (DM), and Natural Language Generation (NLG). Each of these modules can have its own architecture and training regime. For example, a pipeline system could consist of pre-trained neural network NLU and DM components, a rule-based DST, and template NLG. Monolithic systems, on the other hand, tackle the dialog problem at once, modelling the internals of the system as latent states, and only use natural language input-output data pairs for training. Figure~\ref{fig:pipeline_mono} illustrates these two general approaches.
	
	The Eighth Dialog System Technology Challenge \citesystem{DSTC8}{DSTC8} prominently features an \mbox{End-to-End} Multi-Domain Dialog Challenge Track, which requires participants to build a task-oriented agent simulation. The  agent's task is to help the user plan and book a trip around a city, a problem that spans multiple domains ranging from recommending attractions for sight-seeing, to booking transportation (taxi and train) and hotel accommodation. The challenge builds upon two main sources: (i) the MultiWOZ dataset, providing 10,000 human-human interactions in the challenge's target domains, which can be exploited for tasks such as pre-training or system validation; (ii) ConvLab \cite{lee2019convlab}, a novel dialog system framework with special focus on reinforcement learning  (RL) algorithms, enabling direct comparison between the many possible architectures for dialog systems thanks to its \mbox{end-to-end} evaluation. 
	
	For the track, we adopt a pipelined approach to task-oriented dialog systems, and focus on improving the DM component. DM is the decision-making module of the larger dialog system. It uses a policy to choose the next system action given the state of the dialog. The canonical approach is to use rule-based policies, but the problem has also been tackled with supervised and reinforcement learning \cite{wen2016network,li2017end}.
	While rule-based approaches can achieve good performance based on sophisticated, hand-written heuristics, these naturally come at a high development cost: skilled experts are needed to write good rules, and have to stay on hand to extend them to new domains or tasks. 
	On the other hand, RL employs random exploration of a search space to find successful dialog strategies, overcoming the need for hand-crafted rules. However, relying on random discovery of good behaviour can yield limited success in environments where the search space is sufficiently large. For example, in the ConvLab environment provided as part of DSTC8, the state and action spaces are extensive ($2^{392}$ possible states and 300 actions), and the probability of finding 
	paths that lead to successful dialogs through random actions is low.
	
	The contributions presented in this paper are two-fold:\\
	(i) We adapt Deep Q-Learning from Demonstrations \citesystem{DQfD}{DBLP:journals/corr/HesterVPLSPSDOA17}, an RL algorithm that has achieved high scores in Atari game environments, to the dialog domain.
	We employ DQfD to overcome the problem of having to randomly explore a large conversational search space, using its ability to learn to imitate an ``expert'' demonstrator.\\
	(ii) We show that DQfD can be successfully used to train dialog agents
	by leveraging demonstrations that come either from strong rule-based or weaker data-driven experts.
	
	\section{Related Work}
	There is a long history of pipeline approaches to task-oriented dialog systems \cite{levin2000stochastic,walker2000application}. Recently, systems that use neural networks as the basis for one or all of their components have been widely explored \cite{gao2019neural}. 
	Especially for Natural Language Understanding (NLU), pattern recognition techniques such as those involving neural networks are essential. A variety of network architectures have been employed for NLU, including those using conditional random fields \cite{xu2013convolutional}, attention mechanisms \cite{liu2016attention} and Capsule Networks \cite{xia2018zero}.
	
	For Dialog State Tracking and Natural Language Generation components, neural networks save the dialog developer from laboriously writing sets of rules and templates. In the case of DST, approaches range from heuristic trackers -- filling state slots by matching NLU output -- to neural architectures. In particular, long short-term memory (LSTM) networks \cite{zhong2018global} and pointer-networks \cite{WuTradeDST2019} have recently been used to directly learn a mapping from natural language to dialog state.
	Similarly, seq-to-seq models have been applied to NLG to improve the fluency and variety of system utterances over previous approaches
	\cite{wen2015semantically,mei-etal-2016-talk}.
	
	Supervised learning has been used to learn a policy for the Dialog Management module \cite{wen2016network}. But, because DM can be seen as a problem of optimal decision making -- in which past predictions affect future states -- the use of RL is well-established \cite{Young2013POMDPBasedSS}. Recently, neural network based \emph{deep} RL \cite{mnih2015humanlevel} has been used to deal with the combinatorial explosion of possible system states without resorting to feature-engineering \cite{su2017sample}. \shortciteauthor{li2017end} use a deep Q-network (DQN) to model the policy, whereas \shortciteauthor{fatemi2016policy} find that an actor-critic architecture converges more quickly than DQN. 
	
	Despite these improvements in efficiency, using deep RL to learn a dialog policy from scratch is slow, and can fail completely. Training can be bootstrapped by using supervised learning on an in-domain dialog corpus to pre-train the agent's policy \cite{su2017sample}. 
	\shortciteauthor{lipton2018bbq} show that pre-filling the replay buffer with transitions from a rule-based agent improves the chances of DQN converging to a successful dialog policy.
	
	Another barrier to using deep RL for dialog is that the reward is either ill-defined or too sparse. Reward shaping is an active line of research. \shortciteauthor{su2015reward} train an RNN to predict intermediate rewards, whereas \shortciteauthor{takanobu2019guided} use inverse RL to learn a dense reward based on a dialog corpus. Sparse rewards have been tackled in the video games domain. \shortciteauthor{DBLP:journals/corr/HesterVPLSPSDOA17} use expert demonstrations to solve sparse Atari games that had stumped earlier RL approaches while \shortciteauthor{salimans2018learning} learn the notoriously difficult \emph{Montezuma's Revenge} from a single expert demonstration. More efficient exploration has also been shown to improve performance when rewards are sparse \cite{bellemare2016unifying,pathak2017curiosity}. \shortciteauthor{Burda2018ExplorationBR} give the agent an additional reward when it visits unfamiliar states.
	
	\section{Method}
	\label{sec:architecture}
	
	\begin{figure}
		\centering
		\includegraphics[width=.75\columnwidth]{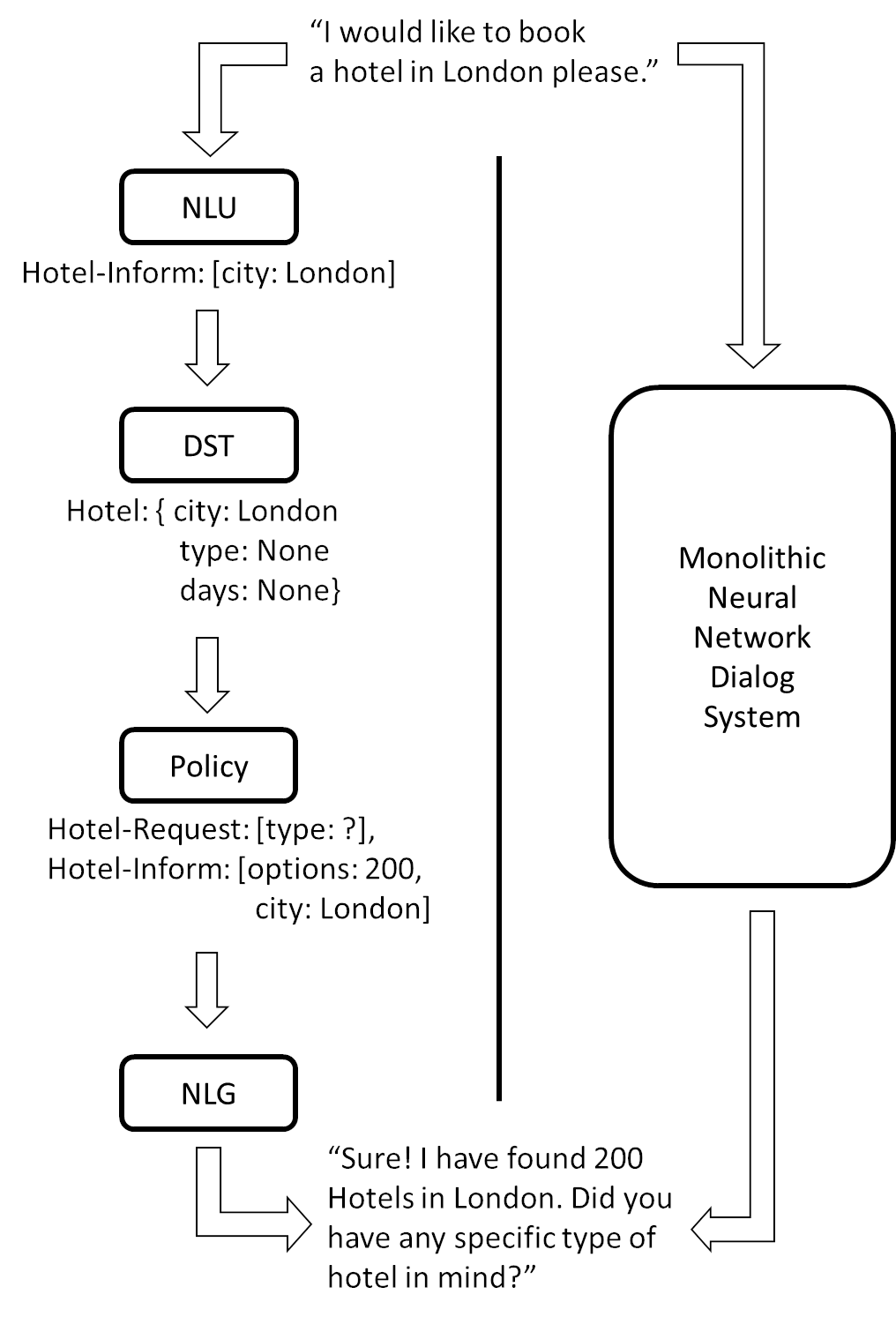}
		\caption{Illustration of pipeline (left) and monolithic neural (right) dialog system architectures.}
		\label{fig:pipeline_mono}
	\end{figure}
	
	For this submission to DSTC8 Track 1, we treat dialog as a modular problem with components specifically designed to handle user input, state tracking, system action prediction, and sentence generation, as illustrated in Figure~\ref{fig:pipeline_mono} (left). This allows for a more direct attribution of changes in system performance to different components than is possible in monolithic systems with latent internal states. We focus our efforts on using RL for the DM module, and keep the other system components fixed, with only minor adjustments to the NLU and NLG modules.
	
	The left-hand side of Figure~\ref{fig:pipeline_mono} outlines the general components of a pipelined \emph{end-to-end} dialog system. End-to-end here refers to the system taking a natural language sentence as input, making decisions based upon this input and current system state, and returning a natural language output sentence. Typically, the individual modules of an end-to-end system internally employ quite different structures for in- and output. In particular, the DM component relies on information being represented in terms of \emph{dialog acts} and system \emph{states}.
	
	Dialog acts (DAs) are abstract representations of information relevant to the dialog. Typically, DAs will encode the dialog \emph{domain} that a particular piece of information is in, as well as associated \emph{slots} and \emph{entities}. For example, the natural language sentence \emph{``I would like to book a hotel in London.''} would be represented by the following DA: \texttt{\{Hotel-Inform: [city: London]\}}.
	
	The dialog \emph{state} is a representation of all the (valid) information collected over the course of a given conversation. Crucially for DSTC8 Track 1, in ConvLab/MultiWOZ notation the state can be sub-divided into information collected by domain, e.g., pertaining to \emph{Hotel} or \emph{Train}.
	Each domain can specify a variety of \emph{slots} that are required to ``complete'' 
	it. For example, the \emph{Train} section of a state may require, amongst others, a date of travel, as well as origin and destination cities.
	
	We describe our approach to learning a dialog policy in \ref{sec:policy}, where we  adapt Deep Q-Learning from Demonstrations (DQfD) to the task-oriented dialog domain. The rest of our proposed architecture, including modules for NLU and NLG, is outlined in \ref{sec:architecture}.
	
	\subsection{Dialog Management}
	\label{sec:policy}
	
	The \emph{policy} of a dialog system is responsible for selecting an appropriate system action, e.g., asking for a specific piece of information, given the current state of the conversation.
	There are many ways to engineer a policy to take the right action, ranging from hand-written rules and heuristics, to machine-learning with data-driven methods or RL. For hand-crafted policies, it can be challenging to specify how a ``good'' system should behave, and it often is time-consuming (and potentially expensive) for human experts to write complex rules to cover all the base- and edge-cases that are needed to achieve the system's desired behaviour. This problem is aggravated in systems that deal with multiple domains at the same time, as is the case for the MultiWOZ-based ``travel agent'' system in ConvLab. Therefore we focus on machine-learning approaches to Dialog Management.
	
	ConvLab offers two ways of training and integrating policies into the dialog system. The first, referred to as ``External Policy'', includes policies that are trained ``outside'' of the full system. This includes rule-based policies (which do not require training), as well as policies trained on in-domain dialog datasets. In terms of neural policy architectures, ConvLab also provides a policy called VMLE (a feed-forward network), as well as an associated ``state featurizer'' used to turn an explicit dialog state into a binary vector representation suitable for input to a neural network model. The second approach to training a policy in ConvLab is to use RL, during which the dialog agent is exposed to a conversation environment in which it interacts with a simulated user trying to achieve a goal, i.e. successfully organizing a city trip.
	
	One potential advantage of RL over supervised learning is that it is able to discover -- and resolve
	-- conversation paths that are rare or, more severely, not even present in available dialog datasets. A well-trained RL agent is therefore likely to outperform its data-driven counterpart, even if only due to the fact that it has been exposed to a larger set of conversational situations.
	
	\subsubsection{Reinforcement Learning}
	In RL for dialog an agent is exposed to a learning \emph{environment} that assigns a reward $r$ to each of its actions. This environment consists of an agenda-based \emph{user-simulator} \cite{schatzmann2007agenda} which typically uses a set of rules to mimic a human user trying to achieve a set of \emph{goals} through talking to the agent. User goals include requesting information (e.g. the price of an attraction) and booking a restaurant, hotel, train or taxi subject to a set of constraints (e.g. number of stars, or train departure time). The more of these goals that the agent satisfies, the higher the reward it receives. The agent does not have access to user goals.
	
	Through interaction with the environment, the agent learns a policy $\pi$ that maps the current state $s_t$ into an action $a_t = \pi(s_t)$. The agent's goal is to maximize its discounted future rewards over the course of a whole dialog. Q-learning \cite{watkins1992q} is a well-established RL algorithm in which the expected total reward of taking an action $a$ in a state $s$ is estimated by the Q-function:
	
	\begin{equation*}
	Q(s, a) = \mathbb{E}_{\pi} \big[ \sum_{k=0}^{T-t} \gamma^{k} r_{t+k}| s_t = s, a_t = a \big]
	\end{equation*}
	
	\begin{equation*}
	\pi^*(s) = \arg \max_a Q^*(s, a)
	\end{equation*}
	
	\vspace*{1ex}
	\noindent where $T$ is the maximum length of the dialog, $t$ is the current dialog turn, and $\gamma$ is a discount factor for future rewards. The goal of Q-learning is to find the \emph{optimal} Q-function $Q^*(s, a)$ with which the expected total reward at each state $s$ is maximized. Given the optimal function $Q^*$, $\pi^*(s)$ is that optimal policy which is obtained by acting greedily in each state according to $Q^*$.
	
	Deep Q-network \citesystem{DQN}{mnih2015humanlevel} is a Q-learning variant that approximates $Q(s, a)$ with a neural network. During training, the agent generates dialogs by acting according to its current policy $\pi$, and stores individual state-action \emph{transitions} in a \emph{replay buffer} in the form ($s_t$, $a_t$, $r_t$, $s_{t+1}$). However, instead of always acting according to $\pi$, an $\epsilon$-greedy strategy is employed in which the agent sometimes takes a random action according to an ``exploration'' parameter $\epsilon$. This forces the agent to diverge from what it currently deems to be the best policy and to explore the state-action search space. Acting $\epsilon$-greedy helps to prevent the agent from getting stuck in local minima by discovering conversation paths with higher total rewards. The transitions aggregated in the replay buffer are then sampled uniformly at regular intervals and used as training examples to update the current estimate of $Q(s, a)$ via the loss:
	\begin{equation*}
	J_{DQ}(Q) = (R(s,a) + \gamma Q(s_{t+1},a_{t+1}^{max};\theta') - Q(s,a;\theta))^2
	\end{equation*}
	
	Through $\epsilon$-greedy exploration, DQN will \emph{eventually} find dialogs that yield a high total reward, but it finds them randomly. This is problematic where the search space is sufficiently large because the agent will likely need to take a large number of random actions to find even a few high return dialogs. This is further aggravated in environments in which rewards are sparse, where the agent has to wait until the end of a conversation to discover whether it has responded to the user appropriately, rather than receiving feedback at every step. It has been shown that otherwise high performing RL algorithms struggle in environments with sparse rewards \cite{Burda2018ExplorationBR}.
	
	The ConvLab environment has both a large state and action space, and sparse rewards. Its default state featurizer converts the explicit dialog state into a 392 dimensional binary vector with $2^{392}$ possible states\footnote{Not all of these states intrinsically ``make sense'', but they can still occur during training as the agent starts out with no knowledge about what constitutes a meaningful state.}; furthermore, there are 300 dialog actions available to the agent at each step in the dialog. Additionally, the agent receives a small penalty to encourage it to minimize dialog length and only receives a large positive or negative reward (depending on whether the user goals were satisfied) at the end of the dialog. Only a small number of randomly chosen state-action combinations will eventually lead to a dialog with a high total reward, so we suspect that $\epsilon$-greedy exploration alone will struggle to converge to good DM performance.
	
	Therefore, we adapt a DQN variant that has been successfully employed to overcome the difficulties of a large state-action space and sparse rewards in a very different domain -- Atari video games.
	
	\subsubsection{Learning Dialog Strategies via Deep Q-Learning from Demonstrations}
	
	We use Deep Q-learning from Demonstrations \citesystem{DQfD}{DBLP:journals/corr/HesterVPLSPSDOA17}, an extension to DQN that uses expert demonstrations to guide the learning process. In this setup, during a pre-training phase a portion of the replay buffer is filled with transitions from an expert demonstrator. The agent learns to imitate these demonstration transitions with the DQN loss function augmented with a large margin classification term:
	
	\begin{equation*}
	J(Q) = J_{DQ}(Q) + \max_{a \in A}[Q(s, a) + l(a_E, a)] - Q(s, a_E)
	\end{equation*}
	
	where $a_E$ is the action the expert demonstrator took in $s$, and $l(a_E, a)$ is 0 when the agent's chosen action is the same as the action taken by the expert demonstrator, and a positive constant $\tau$ otherwise: 
	
	\begin{equation*}
	l(a_E, a)=\begin{cases}
	0, & \text{if $a = a_E$}.\\
	\tau, & \text{otherwise}.
	\end{cases}
	\end{equation*}
	
	After the pre-training phase the agent acts $\epsilon$-greedily according to its current policy and adds its own experience to the replay buffer. By mixing the buffer with demonstration and agent transitions, the agent can learn not just to mimic the expert, but to exceed its performance. Crucially, we keep the portion of the buffer containing expert transitions from being overwritten by newly explored ones. Finally, rather than sampling uniformly from the replay buffer we use a \emph{prioritized replay buffer} to sample in proportion to the training error of a transition \cite{schaul2015prioritized}. We follow \shortciteauthor{DBLP:journals/corr/HesterVPLSPSDOA17} and increase the priority of expert transitions in order to increase their frequency of being sampled during network training. We explore 
	using different experts for DQfD training further in section~\ref{sec:experiments}.
	
	\subsection{System Architecture}
	\label{sec:architecture}
	In addition to a policy, a dialog system requires a number of other key components to fully function. 
	Here we briefly describe the remaining modules we combine with the policy described above, to form our proposed system architecture.
	
	\subsubsection{Natural Language Understanding}
	The first component of our pipelined architecture, the NLU module, utilizes the Multi-intent Language Understanding (MILU) model \cite{lee2019convlab}, one of the reference NLU models provided by ConvLab. MILU is an extension of OneNet \cite{kim2017onenet}, a unified neural network that jointly performs domain, intent, and slot prediction. The original OneNet network has three layers: a character embedding layer which is aggregated using a bidirectional LSTM (BiLSTM); a word-aware layer which combines word embeddings of the aggregated character layer with another BiLSTM; and a multi-output layer which performs domain, intent and slot prediction independently, and is trained by minimizing the joint loss. 
	While OneNet is able to predict a single intent per utterance, MILU represents a multi-intent extension of OneNet with a few additional changes: first, the model replaces the character-based BiLSTM with a CNN layer. It 
	then removes the domain-specific output from OneNet and combines each domain+intent pair into a single output tag. Finally, on top of the word-based BiLSTM, two attention layers are added for both the slot and the domain+intent outputs.
	
	We change the original MILU, replacing the default \mbox{50-dimensional} GloVe word embeddings \cite{pennington2014glove} with contextualized BERT embeddings \cite{devlin2018bert}.
	Specifically we use the cased BERT$_{base}$ model, which consists of a 12-layer Transformer architecture with 768 hidden dimensions and 12 attentions heads. The character encoder remained untouched, employing \mbox{16-dimensional} embeddings combined with a CNN layer with 128 filters of size 3, and ReLU activations.
	
	Table~\ref{tab:nlu} shows the performance in terms of precision, recall, and F1-score of our trained MILU BERT NLU on the MultiWOZ test set, as well as ConvLab's original MILU and OneNet systems. Both MILU-based systems clearly outperform the OneNet NLU, which only achieves around 66\% F1-score on the MultiWOZ test set. The differences between the original MILU and our BERT embedding based system are less pronounced however, our approach achieves higher scores on all metrics.
	
	\begin{table}
		\centering
		\setlength{\tabcolsep}{5pt}
		\begin{tabular}{l|c|c|c|}
			\toprule
			& Prec & Rec & F1 \\
			\midrule
			OneNet & 76.64 & 58.15 & 66.13 \\
			MILU & 76.15 & 81.31 & 78.65 \\
			MILU BERT & \textbf{77.79} & \textbf{81.81} & \textbf{79.75} \\
			\bottomrule
		\end{tabular}
		\caption{Evaluation results of NLU modules trained and tested on MultiWOZ. Reported scores are Precision, Recall, and F1-Score.}
		\label{tab:nlu}
	\end{table}
	
	\subsubsection{Dialog State Tracking}
	The Dialog State Tracker (DST) of a task-oriented dialog system is employed to keep track of the information gathered over the course of a dialog, and unify it with new information coming either from the user via the NLU, or from results of database queries.
	As with other components, there are a variety of ways to derive a DST module including learned, e.g., via data-driven methods, or manually designed rules and heuristics.
	For our experiments in Section~\ref{sec:experiments}, we employ the rule-based DST included in the ConvLab framework. We reason that rule-based state tracking is relatively straightforward\footnote{This is not to say that designing a rule-based DST is a simple task however, we believe it to be more straight-forward than hand-crafting, for example, policy rules or templates for good and diverse natural-language system output.}. For example, a dialog act such as \texttt{\{Hotel-Inform: [city: London]\}} identified by the NLU for the user-uttered sentence \emph{``I would like to book a hotel in London''} can readily be used to fill the \texttt{Hotel} domain's \texttt{city} slot of the dialog state. Similarly, if the DST subsequently queries a hotel database, the query results (e.g., number of available hotels in London) can be added to the state before it is handed over to the policy in order to determine the next system action.
	
	\subsubsection{Natural Language Generation}
	For the NLG component we use a sequence-to-sequence architecture with attention \cite{luong-etal-2015-effective}, to generate natural language sentences from input dialog acts. In order to be robust to unseen words -- which can readily be encountered when having to generate sentences for database entries such as hotel names, or train destinations -- the system is trained on \emph{delexicalized} target sentences. We delexicalize sentences by replacing all values that appear in the source dialog act with placeholder variables.
	For example, in figure~\ref{fig:pipeline_mono} the original target sentence \emph{``I have found 200 hotels
		in London.''} would be changed to \emph{``I have found X-options hotels in X-city.''}. 
	The input to the NLG module is a serialization of the predicate-slot pairs in the dialog act as it is produced by the Dialog Management policy. We do not include the slot values as part of the input to the NLG module, but the variables are replaced with these corresponding values during a post-processing stage to obtain the final output sentence. We implement this architecture by modifying the OpenNMT framework \cite{opennmt}.
	
	The employed attention-based seq-to-seq NLG architectures are prone to producing ``hallucinations'', i.e., they often generate partial output pertaining to slots that are not part of the input dialog act. To ameliorate hallucination in the NLG output, we modify the decoding to never consider output variables that are irrelevant to the input, e.g., we force it to never consider outputting an ``X-price'' variable if the ``price'' slot is not present in the input. Additionally, in order to increase the variety of the output sentences, we use nucleus sampling \cite{DBLP:journals/corr/abs-1904-09751} during decoding.
	
	Table~\ref{tab:nlg} reports BLEU-4 scores \cite{papineni-etal-2002-bleu} achieved on the MultiWOZ test set, for our \mbox{sequence-to-sequence} model as well as ConvLab's template-based NLG, and a semantically-conditioned LSTM \cite{wen2015semantically}. The trainable systems were trained on the MultiWOZ training data. It can be clearly seen that the attention-based seq-2-seq architecture is able to out-perform the two competitor systems in terms of BLEU-4. While both neural systems beat the template-based baseline by a large margin, seq-2-seq further improves upon the SC-LSTM approach by almost 10 BLEU points.
	
	\begin{table}
		\centering
		\setlength{\tabcolsep}{5pt}
		\begin{tabular}{l|c}
			\toprule
			& BLEU-4 \\
			\midrule
			Template & 33.13 \\
			SC-LSTM & 49.01 \\
			Seq-2-Seq Attn & \textbf{58.69} \\
			\bottomrule
		\end{tabular}
		\caption{BLEU-4 results of NLG modules trained and tested on MultiWOZ.}
		\label{tab:nlg}
	\end{table}
	
	\section{Experiments}
	\label{sec:experiments}
	
	\subsection{Experimental Setup}
	We train and evaluate a variety of baselines and target systems, while concentrating heavily on the Dialog Management component. Baselines include a rule-based DM as well as a DQN agent, as provided by ConvLab. We also train a VMLE on the MultiWOZ dataset, for use as system policy. The final trained VMLE achieves accuracy scores of 21.53 (acc@1), 46.96 (acc@3), and 59.06 (acc@5) on the MultiWOZ test set. For DQfD, we experiment with two kinds of experts to provide initial state-action transitions: the rule-based MultiWOZ bot, and an expert using the action predictions of the pre-trained VMLE policy.
	
	We train all RL systems in a DA-to-DA way. In this setup, both the agent and user contain no NLU or NLG components, instead exchanging dialog acts. This effectively makes the system behave as if it was run with ``perfect'' NLU and NLG modules, allowing it to learn a dialog strategy under optimal conditions. We run training or 2.5 million \emph{frames} (user-system turn pairs) with 392-dimensional state feature vector inputs, and 300 target actions. We use $\epsilon$-greedy action exploration and linearly decay $\epsilon$ from 0.1 at the beginning of training to 0.01 after 500k frames. We use a discount factor $\gamma$ of 0.9, and train on 2000 sampled batches of size 32 every 1000 frames. We use a prioritized replay buffer with $\epsilon_p = 0.001$, $\alpha = 0.6$, $\beta = 0.4$ and a maximum buffer size of 100,000 transitions. For DQfD, all transitions generated during pre-training are stored permanently in the buffer, and cannot be overwritten by subsequent agent transitions. All our networks have a single hidden layer of size 100 with a ReLU activation function, have a dueling network structure, and make use of the double DQN loss \cite{van2016deep,wang2015dueling}. We use the RAdam optimizer \cite{liu2019variance} with a learning rate of 0.01, and a step-wise learning rate scheduler. We found that n-step returns do not perform well in the ConvLab environment and drop the n-step term from the DQfD loss as specified by \cite{DBLP:journals/corr/HesterVPLSPSDOA17}. We apply L2 regularization with a weight of $10^{-5}$ and set $\tau$ to a constant 0.8.
	
	\subsection{Results}
	
	We run two sets of evaluation, to assess the agent's learned Dialog Management component in isolation, as well as the full end-to-end system.
	
	In order to evaluate the system's ability to manage the dialog, i.e., react to user inputs and requests with appropriate system actions, we run the first set of experiment in \mbox{DA-to-DA} mode. Since this effectively means that we observe perfect input to the Dialog Management, it means that any performance issues encountered can be attributed to deficits in the management module.
	
	For the second set of experiments we use the NLU and NLG modules as outlined in the previous section, and run the evaluation end-to-end with natural language input and output. As we know how the Dialog Management setups behave in ``perfect'' conditions, any drop in performance in the end-to-end setup is likely to stem from either the system misinterpreting user input, or generating hard to understand output sentences.
	
	For both DA-to-DA and end-to-end setups, all systems were evaluated using the user simulator provided by ConvLab. Evaluation results are shown in Tables~\ref{tab:da2da_eval} and~\ref{tab:e2e_eval_convlab}, for the fully rule-based ``perfect'' baseline (row: \emph{rule}), the MultiWOZ-trained \emph{VMLE} policy, ConvLab's standard \emph{DQN} RL agent, as well as two DQfD variants using a rule-based expert (\emph{Rule DQfD}) and a VMLE-based expert (\emph{VMLE DQfD}) respectively. Results in Table~\ref{tab:da2da_eval} are with disabled NLU and NLG modules to test system performance on ``perfect'' dialog acts. Table~\ref{tab:e2e_eval_convlab} reports results for systems where the user simulator uses the official testing configuration\footnote{according to the task description, the user simulator uses MILU for NLU, and the template-based NLG module for automatic evaluation.}. For RL agents, we evaluate the system at the checkpoint with the best moving average return, over three training sessions. We report the average number of turns, average returned reward, precision, recall, and F1-Score of required user goal slots, as well as percentage  success and book rates.
	
	\subsubsection{DA-to-DA}
	
	\begin{figure}[h!]
		\centering
		\includegraphics[width=.85\columnwidth]{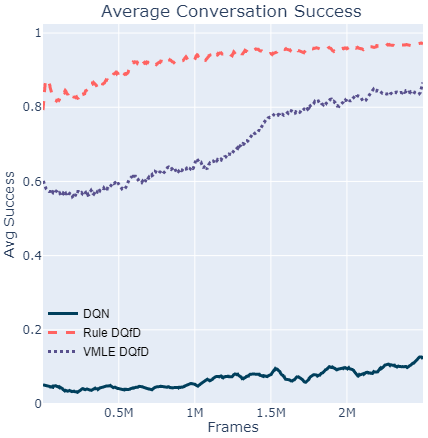}
		\caption{Trendline of system success rates for RL-trained policies, averaged over 3 system training sessions.}
		\label{fig:success}
	\end{figure}
	
	\begin{figure}[h!]
		\centering
		\includegraphics[width=.85\columnwidth]{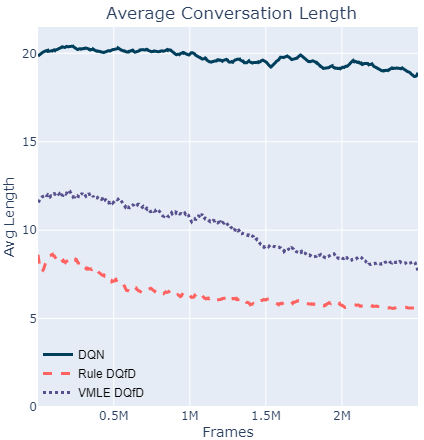}
		\caption{Trendline of system dialog length for RL-trained policies, averaged over 3 system training sessions.}
		\label{fig:length}
	\end{figure}
	
	\begin{table}
		\centering
		\setlength{\tabcolsep}{5pt}
		\begin{tabular}{l||c|c|c|c|c}
			\toprule
			& Rule & VMLE & DQN & DQfD-R & DQfD-V\\
			\midrule
			Turns & 5.25 & 11.67 & 18.79 & 5.33 & 6.81 \\
			Return & 76.75 & 23.53 & -42.57 & 74.67 & 65.19 \\
			\midrule
			Prec & 91.00 & 86.00 & 37.00 & 88.00 & 83.67\\
			Rec & 100 & 80.00 & 26.00 & 99.33 & 91.00\\
			F1 & 94.00 & 81.00 & 29.00 & 92.33 & 85.67\\
			\midrule
			SR & 100 & 61.00 & 15.00 & 98.33 & 85.00\\
			BR & 100 & 52.63 & 15.35 & 97.07 & 88.20\\
			\bottomrule
		\end{tabular}
		\caption{Evaluation results of submitted systems using the ConvLab user simulator, in DA-to-DA setup, for 100 test episodes. Reported scores are average number of dialog Turns, average total Return, Precision, Recall, and F1-Score of required user goal slots, as well as percentage Success and Book Rates (SR and BR).}
		\label{tab:da2da_eval}
	\end{table}
	
	As can be seen from Table~\ref{tab:da2da_eval}, the fully rule-based agent achieves ``perfect'' ceiling performance, with 100\% success rate (SR) and book rate (BR) in the \mbox{DA-to-DA} setup. This serves as a sanity check to ensure the overall system performs as expected. The VMLE network that has been trained on the MultiWOZ dataset shows a significantly worse performance, completing only 61\% of dialogs successfully, while leading to just above half the book-rate of the rule-based agent. Crucially, the bottom three rows of Table~\ref{tab:da2da_eval} show the drastic performance differences between RL agents trained via standard DQN and the proposed DQfD approaches. It can be clearly seen how even when trained \mbox{DA-to-DA}, DQN fails to learn a meaningful policy. After the full 2.5 million frames it only successfully completes 15\% of dialogs. As we hypothesized in section \ref{sec:policy} this is likely a consequence of the large state and action space, and the sparsity of indicative rewards. DQN rarely experiences successful dialogs through exploration. On the other hand, the evaluation results suggest that DQfD successfully overcomes this problem, guiding the agent in the direction of the expert's actions. When the expert demonstrations come from the rule-based expert, the DQfD agent has a set of almost perfect trajectories that it can imitate. Indeed, Table~\ref{tab:da2da_eval} shows that DQfD successfully completes 98.33\% of dialogs and has almost converged to the performance of the rule-based expert. An important finding is the performance of the VMLE-based DQfD. The results show that guidance provided even by this relatively weak expert still allows the DQfD agent to successfully navigate the large state-action space. In fact, when the VMLE is used as a source of demonstrations, the agent does not converge to, but exceeds its performance, achieving a 24\% higher success rate and a 36\% higher book rate than the underlying expert.
	
	All evaluation metrics seem highly correlated, and the same ranking of systems hold for average turn length, average return, success and book rates, as well as most systems' precision, recall and F1-scores. The only exception is the precision achieved by VMLE, which beats the VMLE-based DQfD agent by almost 3\%. However, it in turn suffers from a much lower recall, leading to expected ranking on F1 measure. This correlation of metrics is not very surprising: Precision, recall, and F1-score measure how goot the systems are at fulfilling the user's goal, i.e., leading to correctly filled user goal slots, which is necessarily linked tightly to the overall success and book rates (if slots are not filled/wrongly filled, the user simulator won't book; if slots are correctly filled, it will most likely book). Again closely related is the average return, which necessarily is higher the more often a system is able to successfully manage the dialog, and in such systems that get it mostly right is further influenced by the average number of turns, owed to the constant return ``penality'' of $-1$ per conversation turn. Thus, a system that does very well (or performs very poorly) on either of the evaluated metrics is also highly likely to perform similarly across the others. One notable exception could be the number of turns -- where lower is better -- as a system could feasibly just learn to ``hang up'' on the user in a single turn. However, since the overall training goal is to maximise reward, this is unlikely to happen.
	
	Figures~\ref{fig:success} and~\ref{fig:length} show the trends of average success rate and conversation length the different RL agents achieve over the course of the 2.5 million training frames in the \mbox{DA-to-DA} setting. DQfD gets a head-start over DQN as, during its pre-training phase, it takes actions according to the expert and immediately achieves a relatively high success rate. After pre-training, when the agent acts $\epsilon$-greedily, there is a slight dip in performance as it explores the state-space through random (and probably sub-optimal) actions. This dip can be seen with both rule-based and VMLE experts, but it is only slight. DQfD, encouraged by the large-margin classification term in its loss function, continues to perform similarly to the expert, and eventually converges to, or even exceeds, its success rate as the fraction of random actions decays to a constant 1\%. This is in marked difference to DQN which, without any guidance, languishes at below 10\% for the majority of training time, and only begins to pick up after 1.5 million frames. It might eventually converge if we were to continue training, but it is considerably less sample-efficient than DQfD.
	
	\subsubsection{End-to-End}
	
	\begin{table}[t!]
		\centering
		\setlength{\tabcolsep}{5pt}
		\begin{tabular}{l||c|c|c|c|c}
			\toprule
			& Rule & VMLE & DQN & DQfD-R & DQfD-V\\
			\midrule
			Turns & 8.42 & 12.18 & 18.95 & 9.61 & 11.16\\
			Return & 13.58 & 0.22 & 10.33 & 15.99 & 10.04\\
			\midrule
			Prec & 68.00 & 65.00 & 30.67 & 70.00 & 65.00\\
			Rec & 77.00 & 61.00 & 23.00 & 74.00 & 67.67\\
			F1 & 71.00 & 61.00 & 24.67 & 69.67 & 63.67\\
			\midrule
			SR & 50.00 & 42.00 & 10.33 & 53.00 & 49.33\\
			BR & 86.92 & 62.27 & 19.93 & 75.23 & 69.71\\
			\bottomrule
		\end{tabular}
		\caption{Evaluation results of submitted systems using the ConvLab user simulator, in official evaluation setup, for 100 test episodes. Reported scores are average number of dialog Turns, average total Return, Precision, Recall, and F1-Score of required user goal slots, as well as percentage Success and Book Rates (SR and BR).}
		\label{tab:e2e_eval_convlab}
	\end{table}
	
	\begin{figure}[h!]
		\centering
		\includegraphics[width=\columnwidth]{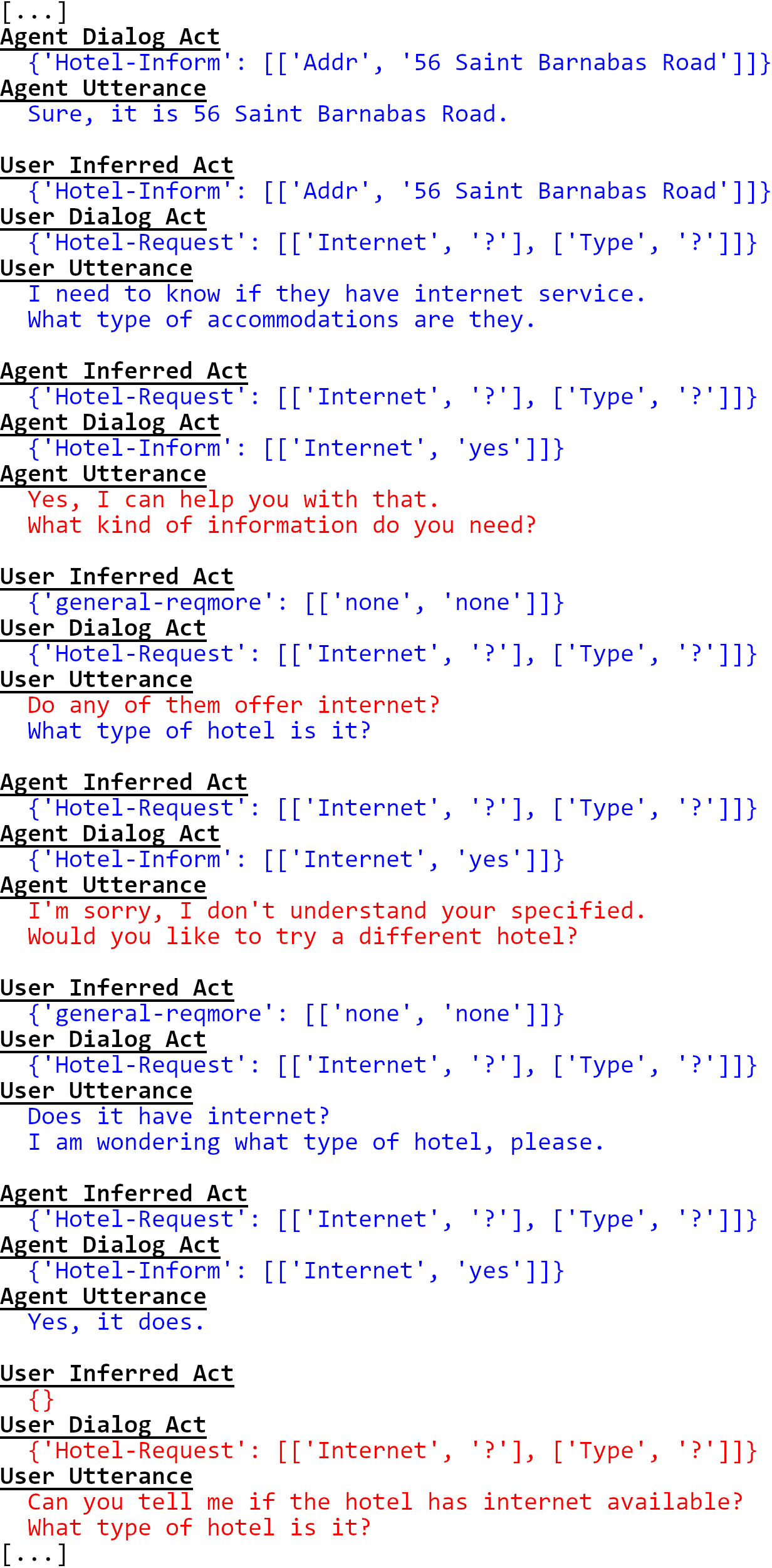}
		\caption{Example of an end-to-end dialog breaking down. \textbf{Inferred Act} denotes what the agent or user NLU understood from the previous \textbf{Utterance}. \textbf{Dialog Act} denotes the actual dialog act intended by the corresponding system, for which its NLG produces the utterance. Instances where NLU/NLG goes right are colored blue, those where some failure occurs are colored red.}
		\label{fig:fail}
	\end{figure}
	
	The systems' performance decreases when they are evaluated end-to-end, as can be seen in Table~\ref{tab:e2e_eval_convlab}. Even with a rule-based DM component, which in isolation satisfies 100\% of the user simulator's goals, the overall system succeeds only 50\% of the time. The overall trend in system performance reflects the same pattern as in the \mbox{DA-to-DA} evaluation. The fully rule-based systems performs best overall, while the agent trained via DQN is unable to find a good dialog strategy. Both DQfD agent variants seem to have learned some good strategies, and the agent trained on VMLE-based expert demonstrations is able to outperform the purely data-driven VMLE policy. Interestingly, the DQfD agent using rule-based demonstrations seems to be able to slightly out-perform the fully rule-based system on certain metrics (precision and book rate). However, overall, it still falls behind on most metrics.
	
	The trend of lower performance in full end-to-end setting holds across all of the configurations that we evaluated. Since we know the systems to perform much better in the DA-to-DA setting -- where it is effectively the DM component that is solely responsible for results -- we attribute this performance drop to errors introduced and propagated by the interactions between the system and user simulator NLU and NLG components. If either side's NLG produces either a wrong utterance, or one that the other system's NLU misrecognizes, the respective dialog state will be updated incorrectly, and subsequently the next performed action will based on incorrect information. This error is compounded if this wrong generation or misunderstanding happens on both sides, and on consecutive turns. Figure~\ref{fig:fail} illustrates a broken-down dialog, where first the agent's NLG fails to produce a valid confirmation sentence to user's request. When it finally manages, after a few re-tries, to produce a proper confirmation sentence, the user-simulator in turn does not understand that its request has just been confirmed. We speculate that system performance in the full end-to-end setting could be improved by either of two means: First, one could target the NLU and NLG systems' weaknesses, and train more reliable modules. The second option would be training the full system in an end-to-end manner, essentially leading to a monolithic system as depicted at the beginning of this paper in Figure~\ref{fig:pipeline_mono}. Either approach comes with its own challenges, however, as we focussed on optimal policy learning, we refer them to future work.
	
	\section{Conclusions and Future Work}
	In this work we have presented our approach to Conversational AI, within the setting of Track 1 of the Eighth Dialog System Technology Challenge (DSTC8).
	We have shown how we adapted and successfully applied \emph{Deep Q-Learning from Demonstrations} (DQfD), originally devised for reinforcement learning in Atari video game environments, to the dialog domain.
	Experiments confirmed the importance of expert demonstrations to RL dialog training. While standard deep reinforcement learning failed to navigate ConvLab's large state and action space and struggled with its sparse rewards, DQfD effectively made use of an expert, which it learned to imitate and eventually outperform. The main advantage of DQfD is that with the expert's help, it can overcome the problem of an RL agent not experiencing vital positive rewards early on in the training process. Furthermore, we have shown that these results hold even when using a relatively weak expert. A DQfD agent using the VMLE expert trained on in-domain data learned to considerably exceed its performance, showcasing the fact that DQfD is not constrained by the quality of the underlying demonstrator.
	
	In future, we would like to further explore the type of ``expert'' needed to successfully train an RL agent via DQfD. While we have shown that an imperfect data-driven expert can be used to train a reliable RL policy, it would be interesting to see how DQfD behaves with even less reliable demonstrations.
	
	\bibliographystyle{dstc8}
	\bibliography{dstc8}
	
\end{document}